\documentclass[11pt,a4paper]{article}
\usepackage[hyperref]{ranlp2023}
\usepackage{times}
\usepackage{latexsym}

\usepackage{microtype}
\usepackage{tabularx}
\usepackage{ragged2e}
\aclfinalcopy 

\title{AHaSIS: Shared Task on Sentiment Analysis for Arabic Dialects}

\author{Maram Alharbi$^1$$^2$, Salmane Chafik$^3$, \textbf{Saad Ezzini$^4$, Ruslan Mitkov$^1$} \\ \textbf{Tharindu Ranasinghe$^1$ and Hansi Hettiarachchi$^1$} \\
 $^1$School of Computing and Communications, Lancaster University, UK \\
 $^2$Jazan University, Saudi Arabia \\
 $^3$Mohammed VI Polytechnic University, Morocco \\
 $^4$King Fahd University of Petroleum and Minerals, Saudi Arabia \\
 {\tt  m.i.alharbi@lancaster.ac.uk} }

\date{}
\usepackage{graphicx}
\usepackage{booktabs}
\usepackage{multirow}
\usepackage{xcolor}

\begin{document}
\maketitle
\begin{abstract}

The hospitality industry in the Arab world increasingly relies on customer feedback to shape services, driving the need for advanced Arabic sentiment analysis tools. To address this challenge, the Sentiment Analysis on Arabic Dialects in the Hospitality Domain shared task focuses on Sentiment Detection in Arabic Dialects. This task leverages a multi-dialect, manually curated dataset derived from hotel reviews originally written in Modern Standard Arabic (MSA) and translated into Saudi and Moroccan (Darija) dialects. The dataset consists of 538 sentiment-balanced reviews spanning positive, neutral, and negative categories. Translations were validated by native speakers to ensure dialectal accuracy and sentiment preservation. This resource supports the development of dialect-aware NLP systems for real-world applications in customer experience analysis. More than 40 teams have registered for the shared task, with 12 submitting systems during the evaluation phase. The top-performing system achieved an F1 score of 0.81, demonstrating the feasibility and ongoing challenges of sentiment analysis across Arabic dialects.

\end{abstract}

\section{Introduction}

Arabic Sentiment Analysis (ASA) has become an increasingly prominent field within Natural Language Processing (NLP), spurred by the growing volume of Arabic content across digital platforms and the pressing need for automated systems to gauge public opinion. In contrast to high-resource languages, ASA continues to face enduring challenges due to the linguistic complexity of Arabic, its diglossic nature, and the considerable variation across regional dialects \cite{Habash2013}. These challenges are particularly evident in informal domains such as social media and hospitality, where sentiment expressions differ significantly across dialects.

To date, the majority of available resources for ASA have concentrated on Modern Standard Arabic (MSA), offering limited applicability to dialectal variants \cite{Aladeemy2024}. Consequently, models trained on MSA frequently struggle to generalise across dialects, leading to diminished performance in practical settings \cite{Khrisat2015}. Additionally, the development of robust, dialect-sensitive models has been hindered by a notable lack of high-quality, annotated datasets.

In response to these limitations, we present the Ahasis 2025 Shared Task, which seeks to advance sentiment classification techniques across Arabic dialects within the hospitality domain. This shared task provides a balanced dataset comprising hotel reviews written in Saudi Arabic and Moroccan Darija, each annotated with sentiment labels. Participants are invited to explore both traditional and neural classification approaches under conditions of limited training data. The task aims to evaluate the effectiveness of various modelling strategies in identifying sentiment from dialect-rich, user-generated content.

The remainder of this paper is organised as follows: Section~\ref{Sec:related_work} reviews the relevant literature; Section~\ref{Sec:task_desc} details the shared task and its setup; Section~\ref{Sec:data} describes the dataset; Section~\ref{Sec:results} presents the evaluation results; and finally, the paper concludes with key findings and outlines future directions.

\section{Related Work} \label{Sec:related_work}
Arabic sentiment analysis has witnessed growing attention in recent years, with early studies laying the foundation by addressing the lack of dialect-specific annotations and lexical resources \cite{Nabil2015}. \citet{Aladeemy2024} critically reviewed the state of sentiment annotation in Arabic dialects, highlighting the prevalence of manual labelling techniques and the limited use of automated methods due to a shortage of robust linguistic resources. Their findings emphasise that machine learning approaches dominate the field, while lexicon-based systems remain underutilised.

Recent literature has placed emphasis on tackling dialectal diversity, recognising that Arabic dialects differ significantly in syntax, morphology, and vocabulary. A systematic review by \citet{Matrane2023} identified key preprocessing stages, such as normalisation, feature extraction, and sentiment tagging, as decisive factors in improving classification performance. The review also underscored the importance of handling negation and morphological variation, both of which are vital to interpreting sentiment in dialectal contexts.

Deep learning architectures, including convolutional neural networks (CNNs) and recurrent models like LSTM \cite{hochreiter1997lstm} and GRU \cite{chung2014gru}, have shown strong results in Arabic sentiment tasks \cite{Baali2019}. However, preprocessing remains a critical bottleneck. \citet{Guellil2020} stressed the necessity of standardised pipelines to improve performance consistency across tasks and datasets.

In parallel, researchers have explored cross-lingual methods to augment Arabic sentiment resources. \citet{Saadany2020} investigated the preservation of sentiment polarity in neural machine-translated Arabic reviews and identified frequent distortions introduced by automated translation tools. Similarly, \citet{Poncelas2020} examined the impact of machine translation on downstream sentiment classification, revealing that models trained on original data outperform those trained on translated corpora, especially in sentiment-sensitive applications.

Finally, while most progress has been made in MSA, \citet{Aladeemy2024} emphasise that Arabic dialects remain underrepresented in sentiment analysis research. They call for a shift towards developing dialect-aware resources and models that address the linguistic variation inherent to Arabic. The Ahasis shared task responds to this call by offering a domain-specific, multi-dialectal dataset and encouraging participants to experiment with resource-efficient and generative learning paradigms.

\section{Task Description}\label{Sec:task_desc}

\subsection{Sentiment Detection in Arabic Dialects}

The Ahasis 2025 Shared Task centres on sentiment analysis within the hospitality domain, specifically targeting hotel reviews written in regional Arabic dialects. Given Arabic’s linguistic richness, marked by the coexistence of MSA and a wide range of spoken dialects, sentiment classification presents notable challenges. Dialects vary considerably in morphology, syntax, and vocabulary, and this variability is further amplified in informal user-generated content, where sentiment is often conveyed through idiomatic or region-specific expressions.

Participants are required to classify hotel reviews into one of three sentiment categories: \textit{positive}, \textit{neutral}, or \textit{negative}. The data comprises user reviews in Saudi and Darija dialects. The task evaluates participants’ ability to build models that can generalise across dialects while maintaining high accuracy in nuanced sentiment interpretation. This shared task explicitly encourages the development of techniques that are resilient to linguistic variation and reflective of real-world text usage in the hospitality sector.

\subsection{Resources and Evaluation}

Participants will be provided with a bi-dialect annotated dataset of hotel reviews. The task permits the use of external resources, including pre-trained encoders, large language models, and data augmentation techniques, allowing for a wide exploration of modelling strategies.

The primary evaluation metric is the \textbf{F1-score}, computed over the three sentiment classes. In addition, secondary analyses will include:
\begin{itemize}
\item \textbf{Dialect-Specific Performance:} Evaluating performance across Saudi and Darija dialects separately.
\item \textbf{Error Categorisation:} Analysing model errors in terms of sentiment misclassification, dialectal confusion, or ambiguous content.
\end{itemize}

\section{Data}\label{Sec:data}

The shared task provides a bi-dialect Arabic sentiment dataset specifically designed for the hospitality domain. The dataset comprises hotel review sentences in two Arabic dialects: Saudi and Moroccan Darija. Each review is annotated with a sentiment label (positive, neutral, or negative), enabling both dialect-specific and cross-dialect sentiment analysis.

The original data was derived from the ABSA-Hotels dataset released as part of the Arabic track of SemEval-2016 \cite{pontiki-etal-2016-semeval, ALDABET2021101224}. This dataset consists of Arabic hotel reviews sourced from platforms such as Booking.com and TripAdvisor. The base data, originally in MSA, was extensively preprocessed and refined following the approach described in \cite{alharbi2025_arabic_llm_sentiment}.

\subsection{Dataset Structure}

The dataset released for this shared task is organized into training and test splits, both covering two Arabic dialects: Saudi and Moroccan Darija. Each instance in the dataset represents a hotel review sentence. The training set includes sentiment annotations, while the test set is used for evaluation and does not expose the sentiment labels.

\begin{table}[ht]
    \centering
    \small
    \begin{tabularx}{\columnwidth}{cccc}
    \hline
    \textbf{Split} & \textbf{Entries} &  \textbf{Dialects} & \textbf{Sentiment Labels} \\
    \hline
    Train & 860 & Saudi, Darija & Positive, Neutral,\\ 
    & & & Negative \\
    Test & 216 & Saudi, Darija & N/A (to be predicted) \\
    \hline
\end{tabularx}
\caption{Structure and statistics of the shared task dataset.}
\label{tab:dataset-structure}
\end{table}

Participants are required to use the provided fields in the test set to predict sentiment labels, ensuring their models generalise well across dialects. The task emphasises robustness to dialectal variation and sentiment nuance, with all reviews grounded in real-world user feedback from the hospitality sector.

\begin{table*}[ht]
    \centering
    \begin{tabular}{lllcc}
    \hline
    \textbf{Submission ID} & \textbf{Codalab Username} & \textbf{Team Name} & \textbf{Test Phase Micro-F1} & \textbf{Rank} \\
    \hline
    281611 & hend\_suliman     & Hend (iWAN-NLP)     & 0.81   & 1 \\
    282197 & ishfmgtun         & ISHFMG\_TUN         & 0.79   & 2 \\
    282404 & nwesri            & LBY                 & 0.79   & 3 \\
    282005 & hasnachouikhi     & LahjaVision         & 0.77   & 4 \\
    282408 & msmadi            & AraNLP              & 0.76   & 5 \\
    282490 & ahmedabdou        & MucAI               & 0.76   & 6 \\
    280604 & shimaa            & MARSAD              & 0.75   & 7 \\
    281739 & almktr            & Lab17               & 0.75   & 8 \\
    281362 & salwas            & BirLee              & 0.75   & 9 \\
    282386 & mabrouka4         & MARSAD AI           & 0.74   & 10 \\
    282374 & mlubbad           & Lubbad              & 0.74   & 11 \\
    282445 & zarnoufi          & MAPROC              & 0.73   & 12 \\
    
    \hline
    \end{tabular}
    \caption{Ahasis Shared Task Test Phase results ranked by Micro-F1.}
    \label{tab:final-results}
\end{table*}
\section{Results and Analysis}\label{Sec:results}

The Ahasis shared task attracted a diverse set of participants, showcasing a range of modelling techniques and domain-specific innovations. Participants engaged with the challenge of accurately classifying sentiment in two dialects, Saudi and Darija, using both fine-tuned transformer models and large language models (LLMs) with prompt engineering strategies.

With sentiment classification task, the competition provided a realistic, low-resource benchmark reflective of linguistic variability and cultural nuance in Arabic dialects. The dataset’s domain focus on hospitality reviews added complexity through indirect sentiment cues, politeness strategies, and dialect-specific idioms. Teams employed diverse strategies, including fine-tuning pretrained models such as MARBERTv2 \cite{abdul-mageed-etal-2021-arbert}, DarijaBERT \cite{gaanoun2024darijabert}, and AraBERT \cite{abdul-mageed-etal-2021-arbert} variants, or leveraging zero-shot capabilities of LLMs like Gemini Pro.

\subsection{Participating Teams and Final Rankings}

A total of 12 teams submitted systems for the test phase. Table~\ref{tab:final-results} presents the final leaderboard based on micro-averaged F1 scores. 

The top-performing system, submitted by Team Hend, achieved an F1 score of 0.81, followed closely by ISHFMG\_TUN and LBY with scores of 0.79.

The results show a tight clustering of top scores between 0.73 and 0.81, with strong performances across a variety of modelling strategies, including fine-tuned transformer models, few-shot LLM prompting, and hybrid lexical-embedding methods.  
\subsection{Team Description}


 \textbf{1. Hend (iWAN-NLP):} The iWAN-NLP team participated in the AHaSIS 2025 shared task with a transformer-based ensemble system designed for sentiment analysis across Arabic dialects. Their approach combined three pre-trained models, MARBERTv2 \cite{abdul-mageed-etal-2021-arbert}, SaudiBERT \cite{qarah2024saudibert}, and DarijaBERT \cite{gaanoun2024darijabert}, each fine-tuned using stratified 5-fold cross-validation. The ensemble was built by averaging logits across folds and models, leveraging model diversity to improve robustness. Training enhancements included label smoothing, mixed-precision training, early stopping, and learning rate warmup. This system achieved a micro F1 score of 0.81, ranking first among all participants.

\noindent \textbf{2. ISHFMG\_TUN:} This team tackled the sentiment analysis task by fine-tuning the AraBERTv02 model \cite{abdul-mageed-etal-2021-arbert}, a pre-trained Arabic language model optimised for social media text. Their approach incorporated several fine-tuning strategies, including freezing lower transformer layers, applying class weighting to address imbalance, and tuning dropout and learning rate schedules. They trained the model using validated on both Saudi and Darija dialects. Without relying on external data, their system achieved a micro F1 score of 0.7916, ranking second in the AHaSIS 2025 shared task.

\noindent \textbf{3. LBY: }The LBY team tackled the AHaSIS 2025 shared task by fine-tuning six pre-trained Arabic transformer models, including bert-base-arabert, bert-base-arabertv02-twitter, bert-large-arabertv02-twitter, MARBERTv2, bert-base-qarib, and DarijaBERT. Their focus was on assessing model performance across both Saudi and Darija dialects. Through a series of dialect-specific and combined-dialect experiments, MARBERTv2 emerged as the top-performing model in their setup. The team emphasised robust training strategies and careful hyperparameter tuning over ensembling, achieving an F1 score of 0.79, and securing third place in the official evaluation. 

\noindent \textbf{4. LahjaVision: } Representing a dialect-focused approach to Arabic sentiment analysis, the LahjaVision team developed a dialect-aware system that leveraged the QARiB transformer model, enriched with specialised dialect embeddings and custom preprocessing for Saudi and Darija Arabic. Their methodology incorporated discriminative fine-tuning, focal loss, and dialect-specific normalisation to better capture sentiment expressions. By embedding dialect information into the model architecture, they achieved notable improvements over baseline and non-dialect-aware systems. Their final system attained a micro F1 score of 0.77, securing fourth place in the AHaSIS 2025 shared task.

\noindent \textbf{5. AraNLP:} Competing in the Ahasis 2025 shared task on sentiment analysis for Arabic hotel reviews, the AraNLP team proposed a hybrid deep learning architecture combining the transformer-based AraELECTRA \cite{antoun-etal-2021-araelectra} model with classical TF-IDF features. This design aimed to capture both contextual semantics and important lexical cues, particularly for dialects like Saudi and Darija. The model was trained with minimal preprocessing to preserve dialectal expressions and used a feature fusion mechanism to integrate embeddings and lexical vectors. AraNLP achieved a micro F1-score of 76\%,  securing fifth place among all participants.

\noindent \textbf{6. MucAI:} The MucAI team approached the AHaSIS 2025 shared task using an innovative few-shot prompting strategy with GPT-4o for Arabic sentiment analysis. They explored zero-shot, static, and adaptive prompting methods, with their final system dynamically retrieving the most semantically similar examples via kNN search over AraBERT embeddings. Each selected example was paired with a chain-of-thought explanation, forming a tailored prompt per review. This adaptive prompting approach significantly improved performance, especially for neutral sentiment cases. The system achieved a micro F1-score of 76\%, outperforming static and zero-shot setups and earning sixth place in the shared task.

\noindent \textbf{7. MARSAD:} This team tackled Arabic sentiment analysis in the hospitality domain by applying structured data augmentation to enhance performance in low-resource dialectal settings. They combined three techniques, paraphrasing via FANAR API, pattern-based sentence generation, and domain-specific word substitution, while retaining dialect-specific linguistic cues. Their approach utilized AraBERT-Large-v02 fine-tuned on both original and augmented data. The resulting system achieved a micro F1-score of 0.75, securing 8th place in the shared task.

\noindent \textbf{8. LAB17:} This team combined generative and transformer-based strategies for sentiment analysis in Arabic dialect hotel reviews. They applied few-shot prompting with GPT-4o and fine-tuned transformer models, including MARBERT and its Omani-dialect variant, SODA-BERT. While GPT-4o reached a micro F1-score of 0.69, their fine-tuned MARBERT model outperformed all with a micro F1-score of 0.75, securing 8th place in the shared task.

\noindent \textbf{9. BirLee:} The BirLee team focused on sentiment analysis for Saudi and Darija dialects by fine-tuning CAMeLBERT-DA with both hotel reviews and a newly curated Saudi proverbs dataset. Their model achieved a micro F1-score of 0.75, outperforming Arabic-centric large language models like Allam 0.70, ACeGPT 0.68, and Jais 0.65 in zero-shot settings. Their results emphasise the effectiveness of domain-specific fine-tuning over zero-shot strategies in dialectal Arabic sentiment analysis.

\noindent \textbf{10. MARSAD AI: } This team tackled sentiment analysis in Arabic dialect hotel reviews through a hybrid model approach. Their system combined contextual embeddings from AraBERT with a custom-built sentiment lexicon tailored to Saudi and Darija dialects. To overcome data scarcity, they implemented two augmentation strategies: probabilistic lexical perturbation and paraphrasing using AraT5. This enriched and diversified the training data. The resulting hybrid model significantly outperformed the baseline AraBERT-only setup, achieving an F1 score of 0.74

\noindent \textbf{11. Lubbad:} Lubbad tackled the sentiment analysis task using the Gemini Pro 1.5 large language model. Instead of retraining, the team employed dialect-specific prompt engineering with real-time batch inference. The approach incorporated sarcasm detection, dialect labelling, and custom zero/few-shot prompts optimized for Saudi and Darija dialects. The system achieved a micro F1-score of 0.7361, ranking 10th in the Ahasis Shared Task.

\noindent \textbf{12. MAPROC:} This team participated in the Ahasis shared task using the SetFit framework, a few-shot learning technique based on fine-tuning sentence transformers. They employed the Arabic-SBERT-100K model and experimented with limited examples per class, ultimately using 64 examples per sentiment category for contrastive fine-tuning. Their approach demonstrated the potential of data-efficient modeling in low-resource dialectal sentiment classification and achieved a micro F1-score of 0.73, placing 12th on the leaderboard.

\section{Conclusion and Future Work}

In conclusion, the Ahasis 2025 Shared Task marks a significant step forward in advancing Arabic sentiment analysis, particularly in addressing the challenges posed by dialectal variation in the hospitality domain. By focusing on sentiment detection in Saudi and Darija dialects, the task has created a valuable benchmark for evaluating NLP systems under low-resource, real-world conditions. The participation of diverse teams employing a range of methodologies, from transformer fine-tuning to few-shot prompting, has yielded meaningful insights into effective modelling strategies for dialectal sentiment classification.

The results highlight the impressive performance of several teams, most notably iWAN-NLP team, whose ensemble of fine-tuned 
BERT-based models achieved the highest F1 score. This underscores the critical role of both model sophistication and dialect-specific data curation in achieving high performance. Moreover, systems that integrated domain knowledge, robust preprocessing, or adaptive prompting techniques also demonstrated strong capabilities, reflecting the importance of combining linguistic insight with technical innovation.

Looking ahead, future work in Arabic dialect sentiment analysis could explore broader dialectal coverage and task extensions such as aspect-based sentiment analysis or emotion detection. Continued development of pre-trained dialectal models and domain-specific embeddings will also be essential for improving generalisability and robustness. 

The Ahasis Shared Task has laid a foundation for future research in Arabic dialect NLP, promoting collaboration and innovation in a field that remains under-represented yet highly impactful. By advancing the development of inclusive, dialect-aware NLP systems, this shared task contributes to broader efforts in enhancing the linguistic diversity, cultural relevance, and real-world applicability of sentiment analysis technologies in the Arab world.

\bibliographystyle{acl_natbib}
\bibliography{ranlp2023}


\end{document}